\documentclass[conference]{IEEEtran}
\IEEEoverridecommandlockouts
\usepackage{cite}
\usepackage{amsmath,amssymb,amsfonts}
\usepackage[greek,english]{babel}
\usepackage[utf8x]{inputenc}
\usepackage{enumitem}
\usepackage{algorithmic}
\usepackage{caption}
\usepackage{subcaption}
\usepackage{graphicx}
\usepackage{textcomp}
\usepackage{xcolor}
\def\BibTeX{{\rm B\kern-.05em{\sc i\kern-.025em b}\kern-.08em
    T\kern-.1667em\lower.7ex\hbox{E}\kern-.125emX}}
\begin{document}

\title{Self-Attention Based Generative Adversarial Networks For Unsupervised Video Summarization\\
}

\author{\IEEEauthorblockN{Maria Nektaria Minaidi}
\IEEEauthorblockA{\textit{School of ECE} \\
\textit{National Technical University of Athens}\\
Athens, Greece \\
minaidimaria@gmail.com}
\and
\IEEEauthorblockN{Charilaos Papaioannou}
\IEEEauthorblockA{\textit{School of ECE} \\
\textit{National Technical University of Athens}\\
Athens, Greece \\
cpapaioan@mail.ntua.gr}
\and
\IEEEauthorblockN{Alexandros Potamianos}
\IEEEauthorblockA{\textit{School of ECE} \\
\textit{National Technical University of Athens}\\
Athens, Greece \\
potam@central.ntua.gr}
}

\maketitle

\begin{abstract}
In this paper, we study the problem of producing a comprehensive video summary following an unsupervised approach that relies on adversarial learning. We build on a popular method where a Generative Adversarial Network (GAN) is trained to create representative summaries, indistinguishable from the originals. 
The introduction of the attention mechanism into the architecture for the selection, encoding and decoding of video frames, shows the efficacy of self-attention and transformer in modeling temporal relationships for video summarization. 
We propose the SUM-GAN-AED model that uses a self-attention mechanism for frame selection, combined with LSTMs for encoding and decoding. 
We evaluate the performance of the SUM-GAN-AED model on the SumMe, TVSum and COGNIMUSE datasets. Experimental results indicate that using a self-attention mechanism as the frame selection mechanism outperforms the state-of-the-art on SumMe and leads to comparable to state-of-the-art performance on TVSum and COGNIMUSE. 
\end{abstract}


\section{Introduction}
\label{sec:intro}
The amount of video data produced on a daily basis is growing at an exponential rate. Given this growth, users increasingly require assistance for selecting, browsing and consuming such extensive collections of videos. Video summarization aims to provide a short visual summary of an original, full-length video, that encapsulates the flow of the story and the most important segments of the video. The goal is for the produced summary to retain only the significant parts and contain as little unnecessary content as possible \cite{survey_apostolidis}. 

Several methods have been proposed to tackle video summarization using information extracted from the audio, video and text modalities.

Early approaches rely on the statistical processing of low-level video, audio and text features for assessing frame similarity or performing clustering-based key-frame selection \cite{Jiang2009}, while the detection of the salient parts of the video is achieved using motion descriptors, color histograms and eigen-features. 

Given the recent growth of neural network architectures, many deep learning based video summarization frameworks have been proposed over the last years. 
Deep-learning based video summarization algorithms typically represent visual content as feature vector encodings of video frames extracted from Convolutional Neural Networks (CNNs)  \cite{survey_apostolidis}. One of the challenges in video summarization is learning the complex temporal dependencies among the video frames. Early temporal modeling approaches used Long Short-Term Memory (LSTM) units \cite{LSTM_first}, or in general, sequence-to-sequence models such as Recurrent Neural Networks (RNNs) \cite{soccer_HMAN,supervised_RNN_zhong}. The introduction of transformers allowed for parallel computation, as well as, better modeling of the long-range temporal dependencies among the video frames \cite{zhao2021hierarchicalTrans}. Generative Adversarial Networks (GANs) \cite{goodfellow2014generative} have also been used in video summarization algorithms. In \cite{Mahasseni_2017_CVPR}, an adversarial framework is proposed consisting of a summarizer and a discriminator, both of which were based on LSTMs. GAN-based video summarization algorithms have been shown to produce state-of-the-art results \cite{survey_apostolidis}. Recently attention mechanisms have appeared to be effective for identifying the important parts of videos \cite{survey_apostolidis, CSNet, SUM-GAN-AAE, conditional}.

In this work, we tackle video summarization as a key-segment selection problem. We adopt a GAN-based video summarization approach and build upon the SUM-GAN model proposed in \cite{Mahasseni_2017_CVPR}. 

Motivated by the limited memorization, long-range and temporal modeling capacity of the LSTM, as well as the success of the multi-head attention and transformer architectures in overcoming these drawbacks \cite{ zhao2021hierarchicalTrans,att-for-vids,trans_korea_multimodal}, we extend SUM-GAN by integrating attention mechanisms in several parts of the architecture. We perform an ablation study to identify the importance of better temporal modeling in the frame selection, encoder and decoder of SUM-GAN. 
We propose the SUM-GAN-AED model that relies on a pure attentional mechanism as the frame selector, while it retains the LSTM module as the encoder and decoder. 

Our contributions include: 1) investigation of the  
efficacy of the integration of transformers into different parts of the SUM-GAN architecture, namely using a transformer for the frame selector (SUM-GAN-SAT), the encoder (SUM-GAN-STD, SUM-GAN-STSED) and the encoder-decoder (SUM-GAN-SAT, SUM-GAN-ST), 2) based on the above results, we propose the use of a self-attention mechanism for the frame selection, while retaining the LSTM architecture for the encoder and the decoder (SUM-GAN-AED), 3) the evaluation of the proposed models on the SumMe, TVSum and COGNIMUSE datasets, achieving state-of-the-art methods performance on SumMe and competitive performance on TVSum and COGNIMUSE. Our experiments indicate that the integration of self-attention at the frame selection stage of the architecture is more effective than at the encoder and decoder stage. Our code implementation can be found at our GitHub Repository\footnote{https://github.com/Aria-Minaidi/Self-Attention-based-GANs-for-Video-Summarization}.

The rest of this paper is outlined as follows: Sections \ref{baseline} and \ref{our model} present the baseline model and the model we propose, respectively. In Section \ref{model variants} we outline the different variants of our proposed model, depending on the position of the added Transformer modules. The experiments are detailed in section \ref{sec:evaluation} and Section \ref{conclusion} concludes the paper.

\section{PROPOSED METHOD}
\label{sec:method}
\subsection{Baseline Model}
\label{baseline}
The basic generative adversarial framework \cite{Mahasseni_2017_CVPR}, is illustrated in Fig. \ref{fig:sumgan}. The frame selector, encoder and decoder jointly constitute the Summarizer, while the decoder (generator) along with the discriminator constitute the Generative Adversarial Network. The encoder and the decoder also form a Variational Auto-Encoder (VAE), which aids training by producing an underlying representation of the video and introducing an additional frame scores vector \cite{vae}. The approach suggests a keyframe selection mechanism, that minimizes the distance between the features of the original videos and the videos that result as a reconstruction from the predicted summaries. The summarizer and the discriminator are trained adversarially in an unsupervised manner, until the discriminator is not able to distinguish between the reconstructed videos from summaries and the original videos.

In more detail, the frame selector is a bi-directional LSTM and the encoder and decoder are LSTMs. Let $\textbf{X} \in \mathbb{R}^{M\times N}$ be the frame features of the input video, derived from the pretrained CNN, where $M$ is the number of frame features and $N$ the number of frames. $\textbf{x}_t \in \mathbb{R}^M$, $t\in [1,N]$, is the feature vector that describes the $t^{th}$ frame. At each instance $t$, the selector is fed with $\textbf{x}_t$ and  outputs a normalized importance scores vector $\textbf{s} \in \mathbb{R}^{N}$, where $\textbf{s}_t \in [0,1]$. The weighted frame features, $\textbf{x}_t\textbf{s}_t$, correspond to the summary and are fed into the encoder which results in a hidden state vector, $\textbf{e} \in \mathbb{R}^H$. The decoder takes $\textbf{e}$ as input and reconstructs a sequence of features representing the input video $\hat{\textbf{x}} \in \mathbb{R}^{N}$. Finally, $\hat{\textbf{x}}$ is forwarded to the LSTM based discriminator, which aims to classify it as 'original' or 'summary'. 

\subsection{The Proposed SUM-GAN-AED Model}
\label{our model}
Our approach introduces an attention-based frame selector in the GAN-based architecture.  The model is inspired by \cite{SUM-GAN-AAE} and \cite{supervised_RNN_zhong}. The self-attention layer ranges over the entire video sequence and efficiently captures long-term temporal dependencies, as opposed to using an LSTM that fails to capture such dependencies effectively \cite{att-for-vids}. This leads to faster computation and a more representative frame selection scores vector \cite{att-for-vids}.

\begin{figure}[t]
    \centering
    \begin{subfigure}{\linewidth}
        \centering 
        \includegraphics[width = 0.7\linewidth]{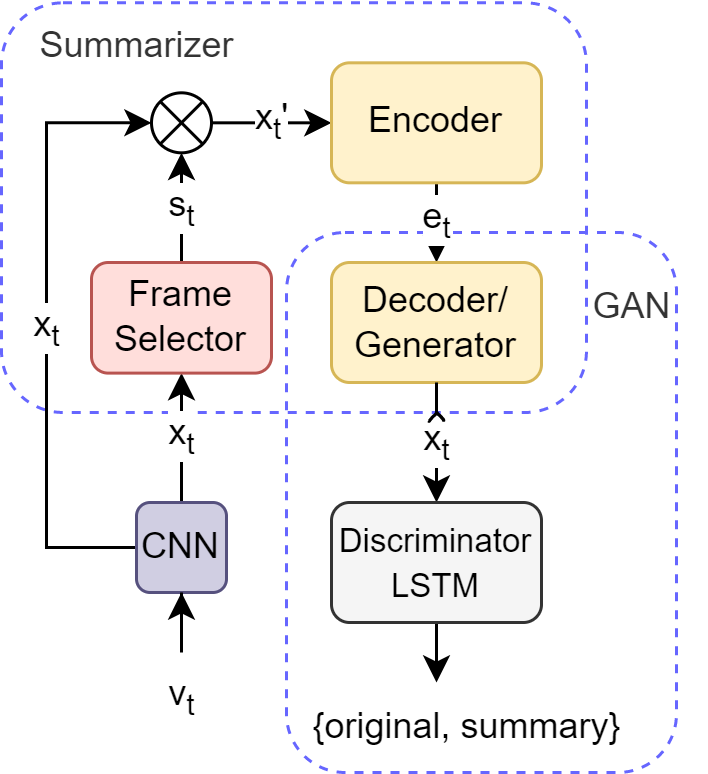}
        \vspace{1.5mm}
        \caption{SUM-GAN}
        \label{fig:sumgan}
    \end{subfigure}
    \hfill
    \vspace{0.2cm}
    \begin{subfigure}{\linewidth}
        \centering 
        \includegraphics[width = 0.7\linewidth]{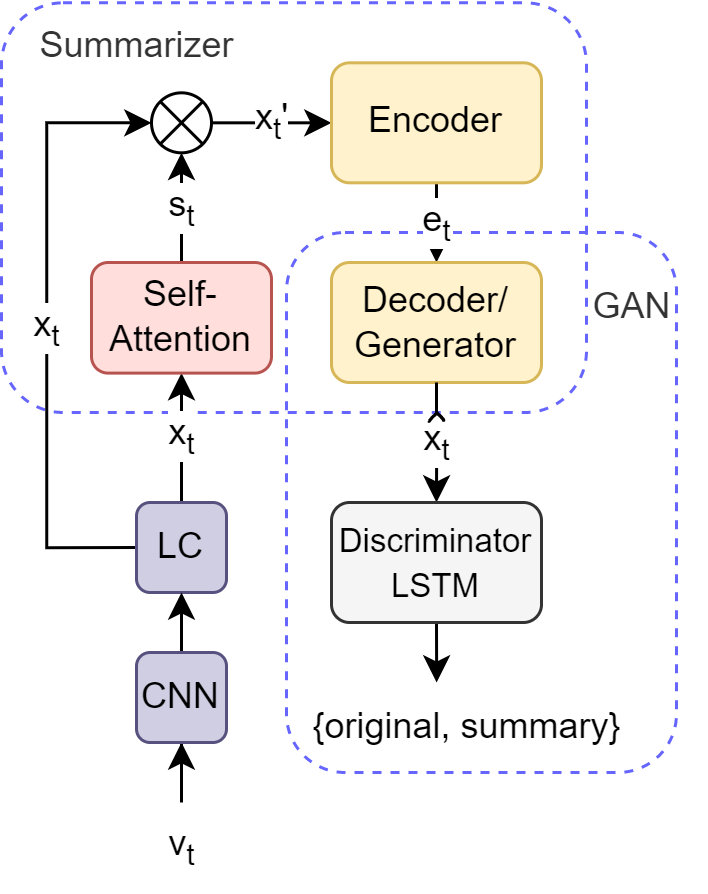}
        \vspace*{-1.5mm}
        \caption{SUM-GAN-AED}
        \label{fig:aed}
    \end{subfigure}
    \caption{The SUM-GAN and SUM-GAN-AED architectures.}
    \label{architectures}
\end{figure}

The architecture of the proposed SUM-GAN-AED model is depicted in Fig. \ref{fig:aed}.
Following the linear compression layer, the compressed frame features sequence is forwarded to the self-attention module. The computed frame scores multiplied with the frame features are then fed into the Variational Auto-Encoder and afterwards to the discriminator. The encoder, decoder and discriminator are LSTMs as in the baseline model.

The parameters of each module of the architecture are learned during training through the loss functions that mirrors \cite{Mahasseni_2017_CVPR} and are as follows:

\begin{itemize}
    \item The Reconstruction Loss: \begin{equation}
        \mathcal{L}_{reconst}=\mathbb{E}[-\log p(\phi(\hat{\textbf{x}})|\textbf{e})]
    \end{equation} 
    where $\phi(\hat{\textbf{x}})$ is the output of the last hidden layer of the discriminator and $p(\phi(\hat{\textbf{x}})|\textbf{e}) \propto exp(-\parallel\phi(\textbf{x}) - \phi(\hat{\textbf{x}})\parallel^2)$.
    
    \vspace{3mm}
    
    \item The Prior Loss:
    \begin{equation}
        \mathcal{L}_{prior}=\mathcal{D}_{KL}(q(\textbf{e}|\textbf{x}) \parallel \mathcal{N}(0,1)))
    \end{equation} where $\mathcal{D}_{KL}$ is the Kullback–Leibler divergence and $q(\textbf{e}|\textbf{x})$ is the probability of observing $\textbf{e}$ given $\textbf{x}$.
    
    \vspace{3mm}
    
    \item The Sparsity Loss: \begin{equation}
        \mathcal{L}_{sparsity}=\parallel\frac{1}{N}\sum_{t=1}^{N}s_t - \text{\foreignlanguage{greek}{σ}} \parallel_2
    \end{equation} 
    where $N$ is the number of video frames and \foreignlanguage{greek}{σ} $\in [0,1]$ is the summary rate that refers to the percentage of frames that constitute the summary. 

    \vspace{3mm}
    
    \item The GAN Loss:
    \begin{equation}
    \begin{aligned}  
        \mathcal{L}_{GAN} = &\log(LSTM(\textbf{x})) + 
        \log(1 - LSTM(\hat{\textbf{x}})) \hspace{1mm} + \\
        &\log(1-LSTM(\hat{\textbf{x}}_p)) 
    \end{aligned}
    \end{equation} where LSTM() is the softmax output of the classifier LSTM and denotes probability scores representing the discriminator’s confidence. $\hat{\textbf{x}}_p$ is
    reconstructed from the subset of video frames randomly selected by sampling from the prior distribution that was produced by the VAE.
    
\end{itemize}

\subsection{Model Variants with Self-Attention}
\label{model variants}

Next, we further investigate the effectiveness of adding self-attention layers in the SUM-GAN model, as it pertains to 
capturing long-term temporal dependencies in video. Specifically,  we experimented with replacing the LSTMs in the SUM-GAN-AED architecture with transformers at various places. The following changes are proposed:
 replacing the LSTM in the encoder with various flavors of a transformer (SUM-GAN-STD, SUM-GAN-STSED) and replacing the LSTMs in the encoder and the decoder with a transformer (SUM-GAN-SAT, SUM-GAN-ST). For the SUM-GAN-STD,  SUM-GAN-STSED and  SUM-GAN-ST models we also use a bi-directional LSTM instead of a self-attention module as the frame selector. The evaluation of the various LSTM, self-attention and transformer variations serve as an ablation study that highlights which modules in the architecture benefit most from better temporal dependency modeling. The proposed models are detailed next.

\vspace{2mm}

1) \textit{\textbf{SUM-GAN-STD}}
\vspace{1mm}\\
In SUM-GAN-STD we swap the encoder LSTM of \cite{Mahasseni_2017_CVPR} for a transformer \cite{attentionisalluneed}. The frame selector is a bi-directional LSTM and the decoder and discriminator are LSTMs as in \cite{Mahasseni_2017_CVPR}. The use of a transformer is inspired by  its ability to mitigate information loss and permit parallel computations \cite{zhao2021hierarchicalTrans, trans_korea_multimodal}. Following the linear compression layer and the frame selector, the weighted feature vector is forwarded to the transformer, whose output is fed to the decoder and then the discriminator. The training follows the baseline \cite{Mahasseni_2017_CVPR}.

\vspace{2mm}

2) \textit{\textbf{SUM-GAN-ST}}
\vspace{1mm}\\
In SUM-GAN-ST we replace the Variational Auto-Encoder, i.e. the encoder and the decoder, with a transformer, in order to enhance the video reconstruction by integrating the positional information of the frames \cite{att-for-vids}. The sequence-to-sequence architecture is suitable for video summarization. The weighted frame features enter the transformer and the reconstructed frame sequence that corresponds to the input video is fed into the discriminator, in order to be classified as 'original' or 'summary'. 
Since we remove the VAE, during training we do not utilize the Prior Loss; the transformer is trained as part of the summarizer and the GAN \cite{SUM-GAN-AAE}.

\vspace{2mm}

3) \textit{\textbf{SUM-GAN-STSED}}
\vspace{1mm}\\
In SUM-GAN-STSED we replace the encoder of the architecture with a \textit{Transformer Sequence Encoder} (TSE), the part of the transformer module that constitutes the encoder \cite{attentionisalluneed}. The intuition remains, a more effective representation of the long-range temporal video dependencies and as a result of the hidden state vector $\textbf{e}$, which leads to a better video reconstruction. For its implementation, we use the slp framework\footnote{https://github.com/georgepar/slp}. The TSE is a sequence-to-vector architecture, which uses positional encodings to insert relative position information of the sequence tokens and keep track of the ordering of the frames.
It takes the output of the frame selector multiplied by the features vector, forwards it to a linear layer, computes the positional embeddings and adds them to the tensor. The resulting vector is forwarded to the encoder part of the TSE and the output is fed to the decoder LSTM of the architecture \cite{att-for-vids}. The frame selector is an LSTM, and the training follows the description in Section \ref{baseline}.

\vspace{2mm}

4) \textit{\textbf{SUM-GAN-SAT}}
\vspace{1mm}\\
Finally, motivated by the effectiveness of the attention based modules, we build SUM-GAN-SAT, in which we swap the LSTM frame selector with a self-attention module and the Variational Auto-Encoder with a transformer. This model combines the SUM-GAN-AED and SUM-GAN-ST architectures and their advantages in integrating attention-based modules in the GAN-based architecture. Here, we incorporate the more thorough frame selection that self-attention provides, with the integration of the positional encodings information during the video reconstruction phase \cite{attentionisalluneed}. We do not use Prior Loss during training, which follows the training of the preceding models.  

\section{EXPERIMENTS}
\label{sec:evaluation}

We evaluate the proposed model against the state-of-the-art on two benchmark datasets SumMe \cite{summe} and TVSum \cite{tvsum}, which contain 25 and 50 videos respectively, with ground-truth annotations. For the ablation study, we add a third database COGNIMUSE \cite{cognimuse} that consists of half-hour segments from seven Hollywood movies. 

\subsection{Evaluation Metrics}
For a fair comparison with the state-of-the-art, we report key-shot-based F-score on TVSum and SumMe datasets, as in \cite{Mahasseni_2017_CVPR}. Specifically the similarity between the automatically generated and the ground-truth summary is computed as the harmonic mean of precision and recall measured on the temporal overlap between the summaries. For the performance evaluation on COGNIMUSE, we follow the approach presented in \cite{koutras}. COGNIMUSE does not include ground-truth frame scores, hence the summarization task is approached as a two-class classification problem, where multiple thresholds are applied to the estimated frame-wise importance scores. Results are obtained at various compression rates, producing summaries of various lengths. The Area Under the Curve (AUC) metric is computed using the Receiver Operating Characteristic (ROC) curve, generated at different thresholds/compression rates.

\subsection{Implementation Details}
For all experiments, we follow the standard 5-fold cross validation approach, 80\% of the videos are used for training and 20\% of the videos are used for testing, and we report the average F-score over the 5 runs. We train our models over 50 epochs.
We downsample the videos to 2 FPS and we use the output of pool5 layer of GoogleNet trained on ImageNet to represent the visual content of the frames. The linear compression layer reduces the size of the feature vectors from 1024 to 500. Each component of the architecture comprises a 2-layer LSTM with 500 hidden units in each layer, while the frame selector LSTM is
bi-directional. Training is based on the Adam optimizer and the learning rate for all components but the discriminator is $10^{-4}$; for the latter it equals to $10^{-5}$. The regularization factor for the sparsity loss is \foreignlanguage{greek}{σ} $ = 0.3$. We implement our method in PyTorch.

\subsection{Results and Comparison}
Table \ref{table:comparison} depicts the F-score results of the recent state-of-the-art unsupervised video summarization methods, in comparison with our model SUM-GAN-AED. The proposed framework surpassed the state-of-the-art methods on SumMe, by a significant margin and on TVSum its performance is comparable to the current state-of-the-art. The efficacy of SUM-GAN-AED is compared to that of the experimental models, developed as part of this study, that utilize transformer architectures in Table \ref{table:results}. Both tables showcase that the introduction of the attention mechanism, either with self-attention or a transformer, leads to an improvement of the performance on SumMe and TVSum, as well as COGNIMUSE.

Between all experimental models, SUM-GAN-AED yields the best overall results, while taking in consideration all the datasets, which also shows that the performance peak is not dataset focused.
The addition of the self-attention mechanism at the frame selection stage in the transformer-based architecture SUM-GAN-ST, leads to improved performance as evidenced by SUM-GAN-SAT. This result underlines the {\em importance of the self-attention module in terms of overall performance}. As it can be seen in Tables \ref{table:comparison} and \ref{table:results}, besides SUM-GAN-AED, our transformer-based models outperform other methods in the literature.

\begin{table}
    \centering
    \caption{Comparative performance evaluation of our SUM-GAN-AED model, with state-of-the-art unsupervised key-frame extraction approaches, on SumMe and TVSum (F-score(\%)).}
        \begin{tabular}{|c|c|c|c|}
        \hline
        \textbf{Model} & \textbf{SumMe} & \textbf{TVSum} \\ \hline
        SUM-GAN \cite{Mahasseni_2017_CVPR} &  38.7 & 50.8  \\ 
        ACGAN \cite{conditional} & 46.0 & 58.5 \\
        Cycle-SUM \cite{cycle_sum} & 46.8 & 57.6 \\
        SUM-GAN-sl \cite{sl} & 46.8  & \textbf{65.3} \\ 
        SUM-GAN-AAE \cite{SUM-GAN-AAE} & 48.9  & 58.3 \\ 
        Proposed-B \cite{icassp} & 58.8 & 63.5 \\ \hline
        SUM-GAN-AED (Ours) &\textbf{64.85}& \textbf{63.18}   \\ 
        \hline        
        \end{tabular}
        \vspace{0.9mm}
        
        \label{table:comparison} 
    \hfill
\end{table}

\begin{table}
    \centering
    \caption{Comparative performance evaluation of our proposed models, on SumMe (F-score(\%)), TVSum (F-score(\%)) and COGNIMUSE (AUC).}
        \begin{tabular}{|c|c|c|c|}
        \hline
        \textbf{Model} & \textbf{SumMe} & \textbf{TVSum} & \textbf{COGN}\\ \hline
        SUM-GAN-STD &  54.15    & \textbf{63.82}   &  51.38   \\ 
        SUM-GAN-ST & 56.00 &   60.53    & 50.73 \\ 
        SUM-GAN-STSED & 61.30 &  62.73 &  52.8    \\ 
        SUM-GAN-SAT & 61.38 &  62.41 & 49.81 \\ \hline
        SUM-GAN-AED &\textbf{64.85}&  \textbf{63.18} &  \textbf{55.49}
        \\ \hline
        \end{tabular}
        \vspace{0.9mm}
        
        \label{table:results} 
    \hfill
\end{table}

\vspace{-0.3mm}
\section{CONCLUSIONS}
\label{conclusion}
We propose a novel framework for unsupervised video summarization based on a Generative Adversarial Network. Building on the SUM-GAN model, we utilize the attention mechanism, with the introduction of self-attention and transformer modules into our framework, in order to capture long-range dependencies, adapt to sequence lengths not encountered in training and embed the positional information of the video frames.
Our experiments show that the use of self-attention for frame selection, followed by LSTMs for encoding and decoding, leads to the best overall results for the video summarization task. Furthermore, the experimental transformer-based architectures perform consistently well on the evaluation datasets and outperform many state-of-the-art unsupervised approaches. Quantitative evaluation on two public benchmark datasets (TVSum, SumMe) and further evaluation on the COGNIMUSE database, proves the efficacy of our approach.

\section*{Acknowledgments}
We would like to thank Georgios Paraskevopoulos and Dimitrios Sotiriou for their valuable conversations and insightful contributions to this research. Their opinions and suggestions have greatly enhanced the quality of this work, and we are thankful for their feedback throughout the process.

\bibliographystyle{IEEEtran}
\bibliography{IEEEabrv,main}

\end{document}